%% file: cvpr20_instance.tex
\documentclass[10pt,twocolumn,letterpaper]{article}

\usepackage{cvpr}
\makeatletter
\@namedef{ver@everyshi.sty}{}
\makeatother
\usepackage{tikz}
\usepackage{times}
\usepackage{epsfig}
\usepackage{graphicx}
\usepackage{amsmath}
\usepackage{amssymb}
\usepackage{comment}
\input{macros}

\usetikzlibrary{decorations.pathreplacing,calc,}
\usetikzlibrary{automata, shapes.geometric,shapes.arrows,decorations.pathmorphing}
\usetikzlibrary{matrix,chains,scopes,positioning,arrows,fit, calc}

\usepackage[pagebackref=true,breaklinks=true,letterpaper=true,colorlinks,bookmarks=false]{hyperref}

\cvprfinalcopy % *** Uncomment this line for the final submission

\def\cvprPaperID{2492} % *** Enter the CVPR Paper ID here

\ifcvprfinal\fi
\begin{document}

%%%%%%%%% TITLE
\title{\papertitle{}}

\author{Francis Engelmann$^{1,2\dagger}$~~~~
Martin Bokeloh$^{2}$~~~~
Alireza Fathi$^{2}$~~~~
Bastian Leibe$^{1}$~~~~
Matthias Nie\ss ner$^{3}$ \vspace{0.1cm}\\
\hspace{-0.5cm} \hfil$^{1}$RWTH Aachen University~~~~~$^{2}$Google~~~~~$^{3}$Technical University Munich \vspace{0.1cm}
}

\twocolumn[{%
\renewcommand\twocolumn[1][]{#1}%
\maketitle
\thispagestyle{empty}
\vspace{-0.8cm}
\includegraphics[width=1.0\linewidth, trim={0 0 0 0cm},clip]{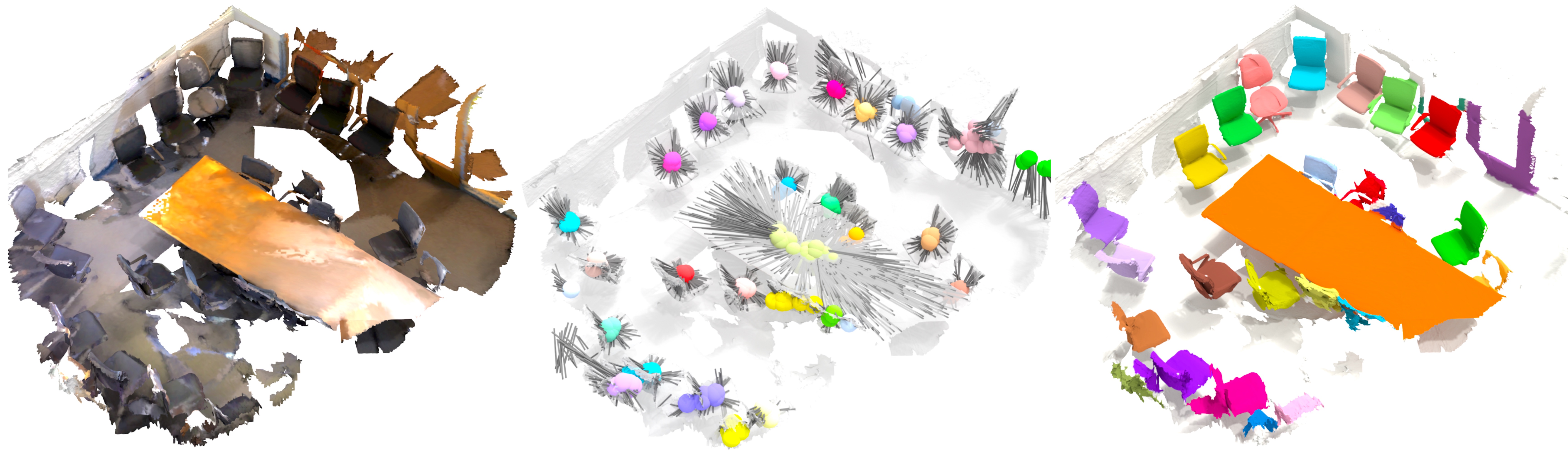}
\begin{tabular}{ccc}
\fcolorbox{gray!0}{gray!0}{\small\textit{Input: 3D Point Cloud }} &
\fcolorbox{gray!0}{gray!0}{\small\textit{Object Center Votes \& Aggregated Proposals}} &
\fcolorbox{gray!0}{gray!0}{\small\textit{Output: 3D Semantic Instances}} \\
\hspace{0.30\textwidth} & \hspace{0.30\textwidth} & \hspace{0.30\textwidth} \\
\end{tabular}
\vspace{-0.7cm}
\captionof{figure}{
Given an input 3D point cloud, our Multi Proposal Aggregation network (3D-MPA) predicts point-accurate 3D semantic instances.
We propose an object-centric approach which generates instance proposals followed by
a graph convolutional network which enables higher-level interactions between adjacent proposals.
Unlike previous methods, the final object instances are obtained by aggregating multiple proposals instead of pruning proposals using non-maximum-suppression.
}
\label{fig:teaser}
\vspace{0.5cm}
}]

\let\thefootnote\relax\footnotetext{$\dagger$ \text{Work performed during internship at Google.}}
\input{sections/0_abstract.tex}
\input{sections/1_introduction.tex}
\input{sections/2_related_work.tex}
\input{sections/3_method.tex}
\input{sections/4_experiments.tex}
\input{sections/5_conclusion.tex}

\balance
{\small
\bibliographystyle{ieee}
\bibliography{abbrev,egbib}
}

\end{document}

%% file: macros.tex
\usepackage{graphicx}
\usepackage{wasysym}
\usepackage{booktabs}
\usepackage{balance}
\usepackage{multirow}
\usepackage{overpic}
\usepackage{bbm}
\usepackage{dsfont}
\usepackage{caption}
\usepackage{color}
\usepackage{pifont}
\usepackage{etoolbox}

\newcommand{\refsec}[1]{Sec.~\ref{sec:#1}}

\newcommand{\reffig}[1]{Fig.~\ref{fig:#1}}
\newcommand{\reftab}[1]{Tab.~\ref{tab:#1}}

\definecolor{m_green}{RGB}{212, 251, 121}
\definecolor{m_orange}{RGB}{255, 212, 121}
\definecolor{m_red}{RGB}{255, 126, 121}
\definecolor{m_violet}{RGB}{215, 131, 255}
\definecolor{m_blue}{RGB}{118, 214, 255}
\newcommand{\colorsquare}[1]{{\color{#1}$\blacksquare$}\hspace{-7.78pt}$\square$}

\newcommand{\circlenum}[1]{{\textcircled{\footnotesize{#1}}}}

\newcommand{\parag}[1]{\vspace{0px}\paragraph{#1}\hspace{-9pt}}

\newcommand{\name}{3D-MPA}
\newcommand{\papertitle}{\name: Multi Proposal Aggregation for 3D Semantic Instance Segmentation}
 
 \definecolor{darkgreen}{RGB}{0,153,51}

%% file: sections/0_abstract.tex
% !TEX root = ../cvpr20_instance.tex
%
\begin{abstract}
We present \name{}, a method for instance segmentation on 3D point clouds.
Given an input point cloud, we propose an object-centric approach where each point votes for its object center.
We sample object proposals from the predicted object centers.
Then, we learn proposal features from grouped point features that voted for the same object center.
A graph convolutional network introduces inter-proposal relations, providing higher-level feature learning in addition to the lower-level point features.
Each proposal comprises a semantic label, a set of associated points over which we define a foreground-background mask, an objectness score and aggregation features.
Previous works usually perform non-maximum-suppression (NMS) over proposals to obtain the final object detections or semantic instances.
However, NMS can discard potentially correct predictions.
Instead, our approach keeps all proposals and groups them together based on the learned aggregation features.
We show that grouping proposals improves over NMS and outperforms previous state-of-the-art methods on the tasks of 3D object detection and semantic instance segmentation on the ScanNetV2 benchmark and the S3DIS dataset.
\end{abstract}

%% file: sections/1_introduction.tex
% !TEX root = ../cvpr20_instance.tex
%
\vspace{-35px}
\section{Introduction}
With the availability of commodity RGB-D sensors such as Kinect or Intel\,RealSense, the computer vision and graphics communities have achieved impressive results on 3D reconstruction methods \cite{Newcombe11ISMAR,Niessner13TOG} that can now even achieve global pose tracking in real time \cite{Dai17TOG,whelan2015elasticfusion}.
In addition to the reconstruction of the geometry, semantic scene understanding is critical to many real-world computer vision applications, including robotics, upcoming applications on mobile devices, or AR/VR headsets.
In order to understand reconstructed 3D environments, researchers have already made significant progress with 3D deep learning methods that operate on volumetric grids \cite{Dai17CVPR,QiCVPR16,Song16CVPR,Song18CVPR,wu20153d}, point clouds \cite{Engelmann20ICRA,Qi17CVPR,Qi17NIPS}, meshes \cite{Hanocka19ACM, Schult20CVPR} or multi-view hybrids \cite{Dai18ECCV,su2015multi}.
While early 3D learning approaches focus mostly on semantic segmentation, we have recently seen many works on 3D semantic instance segmentation \cite{Hou19CVPR,Lahoud19ICCV,Yang19CVPR} and 3D object detection \cite{Qi19ICCV,Zhou18CVPR}, both of which we believe are critical for real-world 3D perception.

One of the fundamental challenges in 3D object detection lies in how to predict and process object proposals:
On one side, top-down methods first predict a large number of rough object bounding box proposals (e.g., anchor mechanisms in Faster R-CNN~\cite{Shaoqing15NIPS}), followed by a second stage refinement step.
Here, results can be generated in a single forward pass, but there is little outlier tolerance to wrongly detected box anchors.
On the other side, bottom-up approaches utilize metric-learning methods with the goal of learning a per-point feature embedding space which is subsequently clustered into object instances~\cite{ElichGCPR19,Lahoud19ICCV,liu2019masc}.
This strategy can effectively handle outliers, but it heavily depends on manually tuning cluster parameters and is inherently expensive to compute at inference time due to $O(N^2)$ pairwise relationships.

In this work, we propose \name{} which follows a hybrid approach that takes advantage of the benefits of both top-down and bottom-up techniques: 
from an input point cloud representing a 3D scan, we generate votes from each point for object centers and group those into object proposals; then -- instead of rejecting proposals using non-maximum-suppression -- we learn higher-level features for each proposal, which we use to cluster the proposals into final object detections. 
The key idea behind this strategy is that the number of generated proposals is orders of magnitude smaller than the number of raw input points in a 3D scan, which makes grouping computationally very efficient.
At the same time, each object can receive multiple proposals, which simplifies proposal generation since objects of all sizes are handled in the same fashion, and we can easily tolerate outlier proposals further down the pipeline.

To this end, our method first generates object-centric proposals using a per-point voting scheme from a sparse volumetric feature backbone.
We then interpret the proposals as nodes of a proposal graph which we feed into a graph convolutional neural network in order to enable higher-order interactions between neighboring proposal features.
In addition to proposal losses, the network is trained with a proxy loss between proposals similar to affinity scores in metric learning;
however, due to the relatively small number of proposals, we can efficiently train the network and cluster proposals.
In the end, each node predicts a semantic class, an object foreground mask, an objectness score, and additional features that are used to group nodes together.

In summary, our contributions are the following:
\begin{itemize}
    \item A new method for 3D instance segmentation based on dense object center prediction leveraging learned semantic features from a sparse volumetric backbone.
    \item To obtain the final object detections and semantic instances from the object proposals, we replace the commonly used NMS with our multi proposal aggregation strategy based on jointly learned proposal features and report significantly improved scores over NMS.
    \item We employ a graph convolutional network that explicitly models higher-order interactions between neighboring proposal features in addition to the lower-level point features.
\end{itemize}

%% file: sections/2_related_work.tex
% !TEX root = ../cvpr20_instance.tex
%
\section{Related Work}
\parag{Object Detection and Instance Segmentation.}
In the 2D domain, object detection and instance segmentation have most notably been influenced by Faster R-CNN from Ren \etal\cite{Shaoqing15NIPS}, which introduced the anchor mechanism to predict proposals with associated objectness scores and regions of interest that enable the regression of semantic bounding boxes.
This approach was extended in Mask-RCNN~\cite{He17ICCV} to predict per-pixel object instance masks.
Hou \etal~\cite{Hou19CVPR} apply the 2D proposal ideas onto the 3D domain by means of dense 3D convolutional networks.
As an alternative, proposal-free methods were proposed in \cite{Brabandere17CVPRW, Fathi17CoRR, Lahoud19ICCV} which rely on metric learning.
In the 2D domain, Fathi \etal~\cite{Fathi17CoRR} estimate how likely pixels are to belong to the same object.
De Brabandere \etal ~\cite{Brabandere17CVPRW} define a discriminative loss, which moves feature points of the same object towards their mean while pushing means of different objects apart.
This discriminative loss is adopted by Lahoud \etal ~\cite{Lahoud19ICCV} to perform instance segmentation in 3D space.
Final instances are obtained via clustering of the learned feature space.
Yang \etal~\cite{Yang19CVPR} directly predict object bounding boxes from a learned global feature vector and obtain instance masks by segmenting points inside a bounding box. 
The recent VoteNet~\cite{Qi19ICCV} highlights the challenge of directly predicting bounding box centers in sparse 3D data as most surface points are far away from object centers.
Instead, they predict bounding boxes by grouping points from the same object based on their votes for object centers.
We adopt the object-centric approach, extend it with a branch for instance mask prediction and replace NMS with a grouping mechanism of jointly-learned proposal features.
\vspace{-10px}
\parag{3D Deep Learning.}
PointNets~\cite{Qi17CVPR} have pioneered the use of deep learning methods for point cloud processing.
Since then, we have seen impressive progress in numerous different fields, including 3D semantic segmentation \cite{Graham18CVPR, Engelmann18ECCVW, Landrieu17CVPR, Qi17CVPR, Qi17NIPS, Tatarchenko18CVPR, Wang18CoRR}, 3D instance segmentation \cite{ElichGCPR19, Hou19CVPR, Lahoud19ICCV, Wang19CVPR, Yang19CVPR, Li19CVPR}, object detection \cite{Hou19CVPR, Qi19ICCV, Zhou18CVPR} and relocalization \cite{Wald19ICCV}, flow estimation \cite{Behl19CVPR, Liu19CVPR, Wang18CVPRa}, scene-graph reconstruction \cite{Armeni19ICCV} and scene over-segmentation \cite{Landrieu19CVPR}.
Point-based architectures, such as PointNet~\cite{Qi19ICCV} and PointNet++~\cite{Qi17ICCV} operate directly on unstructured sets of points, while voxel based approaches, such as 3DMV~\cite{Dai18ECCV} or SparseConvNets~\cite{Choy19CVPR, Graham18CVPR} transform the continuous 3D space into a discrete grid representation and define convolutional operators on the volumetric grid, analogously to image convolutions in the 2D domain. Graph-based approaches \cite{Li16ICLR, Kipf17ICLR, Wang18CoRR} define convolutional operators over graph-structured data such as 3D meshes \cite{Hanocka19ACM, Schult20CVPR}, citation networks \cite{Kipf17ICLR}, or molecules~\cite{Duvenaud15NIPS}.
Here, we leverage the voxel-based approach of Graham \etal~\cite{Graham18CVPR} as point feature backbone and use the graph neural network of Wang \etal~\cite{Wang18CoRR} to enable higher-level interactions between proposals.

\begin{figure*}[th]
\centering

\begin{overpic}[scale=0.59,,tics=2]{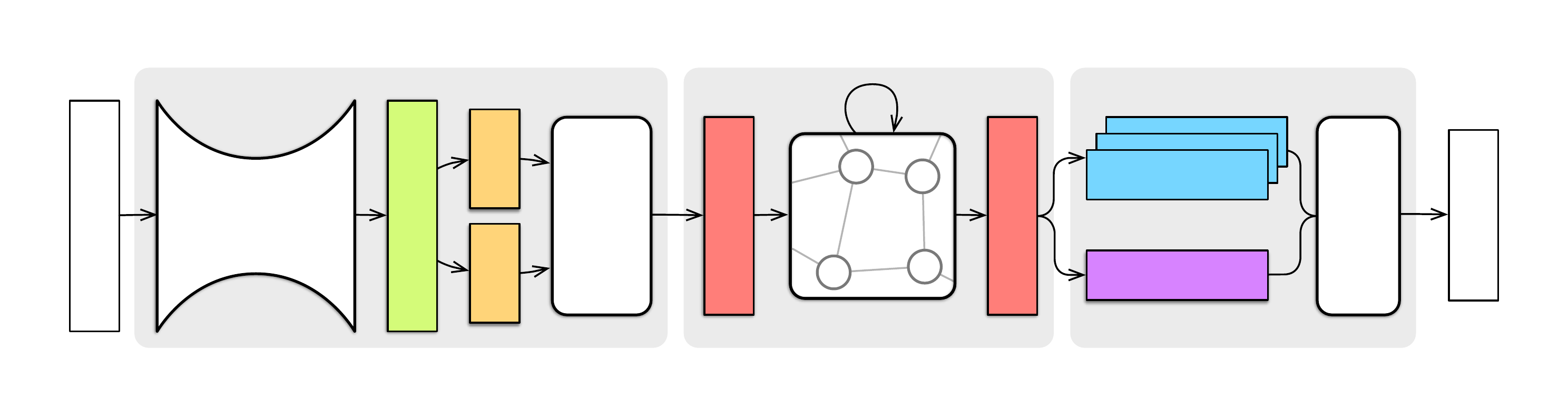}
\put(16,22.7){\textbf{\textit{Proposal Generation}}}
\put(46,22.7){\textbf{\textit{Proposal Consolidation}}}
\put(72,22.7){\textbf{\textit{Object Generation}}}
\put(52.2,10.9){\shortstack[c]{Graph\\ConvNet}}
\put(11.9,10){\shortstack[c]{Sparse\\Volumetric\\Backbone}}

\put(1.5,   0.2){\small\shortstack[c]{Input\\Point Cloud}}
\put(23.5, 0.2){\small\shortstack[c]{Point\\Features}}
\put(43.3, 0.2){\small\shortstack[c]{Proposal\\Features}}
\put(58.6, 0.2){\small\shortstack[c]{Refined Proposal\\Features}}
\put(90.8, 0.2){\small\shortstack[c]{Output\\Objects}}

\put(36.6, 8.2){\rotatebox{90}{\small\shortstack[c]{Sampling \&\\Grouping}}}
\put(85.8, 8.2){\rotatebox{90}{\small\shortstack[c]{Aggregation}}}
\put(93.3, 8.7){\rotatebox{90}{\small\shortstack[c]{K' objects}}}

\put(71, 19.7){\footnotesize{$K$ Proposal Masks}}
\put(29.5, 20.2){\footnotesize{Object Center Votes}}
\put(29.5, 4.6){\footnotesize{Semantic Class}}

\put(5.4, 9.5){\rotatebox{90}{$N \times I$}}
\put(25.6, 9.5){\rotatebox{90}{$N \times F$}}
\put(30.8, 6.1){\rotatebox{90}{$N \times C$}}
\put(30.8, 13.6){\rotatebox{90}{$N \times 3$}}
\put(45.6, 7.2){\rotatebox{90}{$K \times (3+D)$}}
\put(63.7, 7.2){\rotatebox{90}{$K \times (3+D')$}}

\put(72.3, 14.8){$n_i \times 2$}
\put(71.5, 8.3){$K \times D_\text{out}$}
\end{overpic}
\vspace{-0.5cm}
\caption{\textbf{\name{} network architecture}.
From an input point cloud, our network predicts object instance masks by aggregating object proposal masks.
The full model consists of three parts: the proposal generation \emph{(left)} follows an object-centric strategy:
each point votes for the center of the object it belongs to.
Proposal positions are then sampled from the predicted object centers.
By grouping and aggregating votes in the vicinity of sampled proposal positions, we learn proposal features.
During proposal consolidation \emph{(middle)}, proposal features are further refined using a graph convolutional network, which enables higher-order interactions on the level of proposals.
Finally, we propose to aggregate multiple proposals by clustering jointly learned aggregation features as opposed to the commonly used non-maximum-suppression \emph{(right)}.
}
\vspace{-0.3cm}
\label{fig:model}
\end{figure*}

%% file: sections/3_method.tex
% !TEX root = ../cvpr20_instance.tex
%
\section{Method}
The overall architecture of \name{} is depicted in \reffig{model}.
The model consists of three parts:
the first one takes as input a 3D point cloud and learns object proposals from sampled and grouped point features that voted for the same object center (\refsec{proposal_generation}).
The next part consolidates the proposal features using a graph convolutional network enabling higher-level interactions between proposals which results in refined proposal features (\refsec{proposal_consolidation}).
Last, the object generator consumes the object proposals and generates the final object detections, \ie semantic instances.
We parameterize an object as a set of points associated with that object and a semantic class. (\refsec{object_generation}).

\subsection{Proposal Generation}
\label{sec:proposal_generation}
Given a point cloud of size $N \times I$, consisting of $N$ points and $I$-\,dimensional input features (\eg positions, colors and normals),
the first part of the network generates a fixed number $K$ of object proposals.
A proposal is a tuple $(y_i, g_i, s_i)$ consisting of
a position $y_i\in\mathbb{R}^3$,
a proposal features vector $g_i\in\mathbb{R}^{D}$
and a set of points $s_i$ associated with the proposal.

To generate proposals, we need strong point features that encode the semantic context and the geometry of the underlying scene.
We implement a sparse volumetric network \cite{Choy19CVPR, Graham18CVPR} as feature backbone to generate per-point features $\{f_i \in \mathbb{R}^F \}_{i=1}^{N}$ (\reffig{model}, \colorsquare{m_green}).
Semantic context is encoded into the point features by supervising the feature backbone with semantic labels,
using the standard cross-entropy loss for per-point semantic classification $\mathcal{L}_\text{sem.pt.}$.
Following the object-centric approach suggested by Qi \etal.~\cite{Qi19ICCV}, points vote for the center of the object they belong to.
However, unlike \cite{Qi19ICCV}, only points from objects predict a center.
This is possible since we jointly predict semantic classes, \ie we can differentiate between points from foreground (objects) and background (walls, floor, \etc) during both training and test.
This results in precise center predictions since noisy predictions from background points are ignored.
In particular, this is implemented as a regression loss which predicts per-point  relative 3D offsets $\Delta x_i \in \mathbb{R}^3$ between a point position $x_i \in \mathbb{R}^3$ and its corresponding ground truth bounding-box center $c_i^* \in \mathbb{R}^3$. We define the per-point center regression loss as:
\begin{equation}
	\mathcal{L}_\text{cent.pt.}=\frac{1}{M}|| x_i + \Delta x_i - c_i^* ||_H \cdot \mathds{1}(x_i) ~,
\end{equation}
where $||\cdot||_H$ is the \emph{Huber}-loss (or smooth L$_1$-loss) and $\mathds{1}(\cdot)$ is a binary function indicating whether a point $x_i$ belongs to an object. $M$ is a normalization factor equal to the total number of points on objects.
All in all, the feature backbone has two heads (\reffig{model}, \colorsquare{m_orange}): a semantic head (which performs semantic classification of points) and a center head (which regresses object centers for each point). They are jointly supervised using the combined loss  $\mathcal{L}_\text{point}$ where $\lambda$ is a weighting factor set to 0.1:
\begin{equation}
    \mathcal{L}_\text{point} = \lambda\cdot\mathcal{L}_\text{sem.pt.} + \mathcal{L}_\text{cent.pt.}.
\end{equation}

\vspace{-15px}
\parag{Proposal Positions and Features.}
After each point (that belongs to an abject) has voted for a center,
we obtain a distribution over object centers (\reffig{qualitative_results}, $3^\text{rd}$ col.).
From this distribution, we randomly pick $K$ samples as proposal \emph{positions} $\{y_i = x_i + \Delta x_i \in \mathbb{R}^3 \}_{i=1}^{K}$ (\reffig{qualitative_results},  $4^\text{th}$ col.).
We found random sampling to work better than \emph{Farthest Point Sampling} (FPS) used in \cite{Qi19ICCV},
as FPS favors outliers far away from true object centers.
Next, we define the set of associated points $s_i$  as those points that voted for centers within a radius $r$ of the sampled proposal position $y_i$.
The proposal \emph{features} $\{g_i \in \mathbb{R}^D \}_{i=1}^{K}$ are learned using a PointNet~\cite{Qi17CVPR} applied to the point features
of the associated points $s_i$.
This corresponds to the grouping and normalization technique described in \cite{Qi19ICCV}.
At this stage, we have $K$ proposals composed of 3D positions $y_i$ located near object centers, proposal features $g_i \in \mathbb{R}^D$ describing the local geometry and the semantics of the nearest objects (\reffig{model}, \colorsquare{m_red}), along with a set of points $s_i$ associated with each proposal.

\subsection{Proposal Consolidation}
\label{sec:proposal_consolidation}
\vspace{-5px}
So far, proposal features encode \emph{local} information of their associated objects.
During proposal consolidation, proposals become aware of their \emph{global} neighborhood by explicitly modeling higher-order interactions between neighboring proposals.
To this end, we define a \emph{graph convolutional network} (GCN) over the proposals.
While the initial point-feature backbone operates at the level of points, the GCN operates at the level of proposals.
In particular, the nodes of the graph are defined by the proposal positions $y_i$ with associated proposal features $g_i$.
An edge between two nodes exists if the Euclidean distance $d$ between two 3D proposal positions $y_{\{i,j\}}$ is below 2\,m.
We adopt the convolutional operator from DGCNN~\cite{Wang18CoRR} to define edge-features $e_{ij}$ between two neighboring proposals as:
\begin{equation}
    e_{ij} = h_\Theta\big([y_i, g_i], [y_j, g_j] - [y_i, g_i]\big) ~,
\end{equation}
where $h_\Theta$ is a non-linear function with learnable parameters $\theta$ and $[\cdot, \cdot]$ denotes concatenation. The graph convolutional network consists of $l$ stacked graph convolutional layers.
While our method also works without the GCN refinement (\ie $l$\,=\,$0$), we observe the best results using $l$\,=\,$10$ (\refsec{results}).
To conclude, during proposal consolidation a GCN learns refined proposal features $\{h_i \in \mathbb{R}^{D'} \}_{i=1}^{K}$ given the initial proposal features $\{g_i \in \mathbb{R}^{D} \}_{i=1}^{K}$ (\reffig{model}, \colorsquare{m_red}).

\subsection{Object Generation}
\label{sec:object_generation}
\vspace{-5px}
At this stage, we have $K$ proposals $\{(y_i, h_i, s_i)\}_{i=1}^K$ with positions $y_i$, refined features $h_i$ and sets of points $s_i$.
The goal is to obtain the final semantic instances (or object detections) from these proposals.
To this end, we predict for every proposal a semantic class, an aggregation feature vector, an objectness score and a binary foreground-background mask over the points $s_i$ associated with the proposal.
Specifically, the proposal features $h_i$ are input to an MLP with output sizes $(128, 128, D_{out})$
where $D_{out} = S + E + 2$ with $S$ semantic classes,
$E$-dimensional aggregation feature and a 2D (positive, negative) objectness score (\reffig{model}, \colorsquare{m_violet}).

The objectness score \cite{Qi19ICCV,Shaoqing15NIPS} classifies proposals into positive or negative examples.
It is supervised via a cross-entropy loss $\mathcal{L}_{obj.}$.
Proposals near a ground truth center ($< 0.3$~m) are classified as positive.
They are classified as negative, if they are far away ($> 0.6$~m) from any ground truth center,
or if they are equally far away from two ground truth centers since then the correct ground truth object is ambiguous.
This is the case when $d_1 > 0.6\cdot d_2$ where $d_i$ is the distance to the $i^{th}$ closest ground truth center.

Positive proposals are further supervised to predict a semantic class, aggregation features, and a binary mask.
Negative ones are ignored.
We use a cross-entropy loss $\mathcal{L}_\text{sem.}$ to predict the semantic label of the closest ground truth object.

\vspace{-13px}
\parag{Aggregation Features.}
Previous methods such as VoteNet\,\cite{Qi19ICCV} or 3D-BoNet\,\cite{Yang19CVPR} rely on non-maximum-suppression (NMS) to obtain the final objects.
NMS iteratively selects proposals with the highest objectness score and removes all others that overlap with a certain IoU.
However, this is sensitive to the quality of the objectness scores and can discard correct predictions.
Instead of rejecting potentially useful information, we combine multiple proposals.
To this end, we learn aggregation features for each proposal which are then clustered using DBScan\,\cite{Ester96KDD}.

All proposals whose aggregation features end up in the same cluster are aggregated together, yielding the final object detections. 
The points of a final object are the union over the foreground masks of combined proposals.
As the number of proposals is relatively small ($K$\,$\approx$\,$500$) compared to the full point cloud ($N$\,$\approx$\,$10^6$), this step is very fast ($\sim 8$\,ms).
This is a significant advantage over clustering full point clouds \cite{ElichGCPR19, Lahoud19ICCV}, which can be prohibitively slow.

We investigate two types of aggregation features:\\
\circlenum{1} \textit{Geometric features} $\{\epsilon_i \in \mathbb{R}^{E=4} \}_{i=1}^{K}$ are composed of a refined 3D object center prediction $\Delta y_i$ and a 1D object radius estimation $r_i$.
The loss is defined as: 
\vspace{-3px}
\begin{equation}
    \mathcal{L}_\text{agg.} = ||y_i + \Delta y_i - c_i^* ||_H + ||r_i - r_i^*||_H
\vspace{-3px}
\end{equation}
where $c_i^*$ is the nearest ground truth object center and $r_i^*$ the radius of the nearest ground truth object bounding sphere.\\
\circlenum{2} \textit{Embedding features} $\{\epsilon_i \in \mathbb{R}^E \}_{i=1}^{K}$ are supervised with a discriminative loss function\,\cite{Brabandere17CVPRW}.
This loss was already successfully applied for 3D instance segmentation \cite{ElichGCPR19,Lahoud19ICCV}.
It is composed of three terms:
$\mathcal{L}_\text{agg.} = \mathcal{L}_\text{var.} + \mathcal{L}_\text{dist.}  + \gamma \cdot \mathcal{L}_\text{reg.}$
\begin{equation}
    \mathcal{L}_\text{var.} = \frac{1}{C}\sum_{c=1}^{C}\frac{1}{N_C}\sum_{i=1}^{N_C}\,[||\mu_C - \epsilon_i|| - \delta_v]_+^2
\end{equation}
\begin{equation}
    \mathcal{L}_\text{dist.} = \frac{1}{C(C-1)}\underset{C_A \neq C_B}{\sum_{C_A=1}^C \sum_{C_B=1}^C} [2\delta_d - ||\mu_{C_A} - \mu_{C_B}||]^2_+
\end{equation}
\begin{equation}
    \mathcal{L}_\text{reg.} = \frac{1}{C}\sum_{C=1}^C||\mu_C||
\end{equation}
In our experiments, we set $\gamma = 0.001$ and $\delta_v = \delta_d = 0.1$.
$C$ is the total number of ground truth objects and $N_C$ the number of proposals belonging to one object.
$\mathcal{L}_\text{var.}$ pulls features that belong to the same instance towards their mean,
$\mathcal{L}_\text{dist.}$ pushes clusters with different instance labels apart,
 and $\mathcal{L}_\text{reg.}$ is a regularization term pulling the means towards the origin.
 Further details and intuitions are available in the original work by DeBrabandere \etal~\cite{Brabandere17CVPRW}.
In \refsec{results}, we will show that geometric features outperform embedding features.

\vspace{-13px}
\parag{Mask Prediction.}
Each positive proposal predicts a class-agnostic binary segmentation mask over the points $s_i$ associated with that proposal, where the number of points per proposal $i$ is $|s_i| = n_i$ (\reffig{model}, \colorsquare{m_blue}).
Prior approaches obtain masks by segmenting 2D \emph{regions of interest} (RoI) (Mask-RCNN~\cite{He17ICCV}) or 3D bounding boxes (3D-BoNet~\cite{Yang19CVPR}). 
Since we adopt an object-centric approach, mask segmentation can directly be performed on the points $s_i$ associated with a proposal.
In particular, for each proposal, we select the per-point features $f_i$ of points that voted for a center within a distance $r$ of the proposal position $y_i$.
Formally, the set of selected per-point features is defined as $M_{f} = \{f_i |\, \lVert(x_i + \Delta x_i) - y_i\rVert_2 < r\}$ with $r = 0.3$\,m.
The selected features $M_{f}$ are passed to a PointNet~\cite{QiCVPR16} for binary segmentation, \ie, we apply a shared MLP on each per-point feature, compute max-pooling over all feature channels, and concatenate the result to each feature before passing it through another MLP with feature sizes (256, 128, 64, 32, 2).
Points that have the same ground truth instance label as the closest ground truth object instance label are supervised as foreground, while all others are background.
Similar to \cite{Yang19CVPR}, the mask loss $\mathcal{L}_\text{mask}$ is implemented as \emph{FocalLoss}~\cite{Lin17ICCV} instead of a cross-entropy loss to cope with the foreground-background class imbalance.

\subsection{Training Details}
\vspace{-5px}
The model is trained end-to-end from scratch using the multi-task loss
$\mathcal{L} = \mathcal{L}_\text{point} + \mathcal{L}_\text{obj.} + 0.1\cdot\mathcal{L}_\text{sem.} + \mathcal{L}_\text{mask} + \mathcal{L}_\text{agg.}$. 
The batch size is 4 and the initial learning rate 0.1 which is reduced by half every 2$\cdot10^4$ iterations and trained for 15$\cdot10^4$ iterations in total.
Our model is implemented in TensorFlow and runs on an Nvidia TitanXp GPU (12GB). 
\vspace{-13px}
\parag{Input and data augmentation.}
Our network is trained on 3\,m$\times$3\,m point cloud crops of $N$ points sampled from the surface of a 3D mesh.
During test time, we evaluate on full scenes.
Input features are the 3D position, color and normal assigned to each point. 
Data augmentation is performed by randomly rotating the scene by $\text{Uniform}[-180^{\circ}, 180^{\circ}]$ around the upright axis and $\text{Uniform}[-10^{\circ}, 10^{\circ}]$ around the other axis. The scenes are randomly flipped in both horizontal directions and randomly scaled by $\text{Uniform}[0.9, 1.1]$.

%% file: sections/4_experiments.tex
% !TEX root = ../cvpr20_instance.tex

\section{Experiments}
\label{sec:results}
\vspace{-5px}
We compare our approach to previous state-of-the-art methods on two large-scale 3D indoor datasets (\refsec{comparison}). 
Our ablation study analyzes the contribution of each component of our model and shows in particular the improvement of aggregating proposals over NMS (\refsec{ablation}). 

\input{results/object_detection_scannet.tex}
\input{results/instance_segmentation_stanford.tex}
\input{results/instance_segmentation_scannet.tex}

\input{results/qualitative_results.tex}

\input{results/instance_segmentation_scannet_class.tex}

\subsection{Comparison with State-of-the-art Methods}
\label{sec:comparison}

\parag{Datasets.} The \emph{ScanNetV2}~\cite{Dai17CVPR} benchmark dataset consists of richly-annotated 3D reconstructions of indoor scenes. It comprises 1201 training scenes, 312 validation scenes and 100 hidden test scenes. The benchmark is evaluated on 20 semantic classes which include 18 different object classes.

The \emph{S3DIS}~\cite{Armeni16CVPR} dataset is a collection of six large-scale indoor areas annotated with 13 semantic classes and object instance labels. We follow the standard evaluation protocol and report scores on Area 5, as well as 6-fold cross validation results over all six areas.

\vspace{-5px}
\parag{Object detection scores} are shown in \reftab{object_detection_scannet}.
Object detections are obtained by fitting a tight axis-aligned bounding box around the predicted object point-masks.
We compare \name{} to recent approaches including \emph{VoteNet}~\cite{Qi19ICCV} on the ScanNetV2~\cite{Dai17CVPR} dataset. Scores are obtained by using the evaluation methodology provided by \cite{Qi19ICCV}.
Our method outperforms all previous methods by at least \textbf{+5.8 mAP@25\%} and \textbf{+15.7 mAP@50\%}.
\input{results/ablation_study.tex}

\input{results/instance_segmentation_stanford_class.tex}

\vspace{-5px}
\parag{Instance segmentation scores} on S3DIS~\cite{Armeni16CVPR} are shown in \reftab{inst_stan}.
Per-class instance segmentation results are shown in \reftab{instance_segmentation_stanford_class}.
We report mean average precision (mAP) and mean average recall (mAR) scores.
Our scores are computed using the evaluation scripts provided by Yang \etal~\cite{Yang19CVPR}.
Our approach outperforms all previous methods. In particular, we report an increased recall of \textbf{+17.8 mAR@50\%} on Area5 and \textbf{+16.5 mAR@50\%} on 6-fold cross validation, which means we detect significantly more objects, while simultaneously achieving higher precision. 
\newpage
We show results on ScanNetV2~\cite{Dai17CVPR} validation and hidden test set in \reftab{inst_scan} and per-class scores with mAP@25\% in \reftab{instance_segmentation_scannet_class_map25} and mAP@50\% in \reftab{instance_segmentation_scannet_class_map50}.
We improve over previous methods by at least \textbf{+18.1 mAP@50\%} and \textbf{+17.0 mAP@25\%}.
In particular, our \name{} outperforms all other methods in every object class on mAP@50 (\reftab{instance_segmentation_scannet_class_map50}). On mAP@25,  we outperform on all classes except \emph{chair} and \emph{sofa}.
Qualitative results on ScanNetV2 are visualized in \reffig{qualitative_results} and failure cases in \reffig{failure_cases}.

\subsection{Ablation study}
\label{sec:ablation}
In \reftab{ablation_scannet}, we show the result of our ablation study analyzing the design choices of each component of our model.
The evaluation metric is mean average precision (mAP) on the task of instance segmentation, evaluated on the ScanNetV2 validation set.

\parag{Effect of grouping compared to NMS.}
The main result of this work is that grouping multiple proposals is superior to non-maximum-suppression (NMS).
We demonstrate this experimentally by comparing two baseline variants of our model:
In experiment \circlenum{1} (\reftab{ablation_scannet}), we apply the traditional approach of predicting a number of proposals and applying NMS to obtain the final predictions. The model corresponds to the one depicted in \reffig{model} without proposal consolidation and with the aggregation replaced by NMS.
NMS chooses the most confident prediction and suppresses all other predictions with an IoU larger than a specified threshold, in our case 25\%. 
For experiment \circlenum{2}, we perform a naive grouping of proposals by clustering the proposal positions $y_i$.
The final object instance masks are obtained as the union over all proposal masks in one cluster.
We observe a significant increase of \textbf{+4.9 mAP} by replacing NMS with aggregation.

\parag{How important are good aggregation features?} In experiment \circlenum{2}, we group proposals based on their position $y_i$. These are still relatively simple features. In experiments \circlenum{3} and \circlenum{4}, proposals are grouped based on learned embedding features and learned geometric features, respectively. These features are described in \refsec{object_generation}. Again, we observe a notable improvement of +5.4 mAP compared to experiment \circlenum{2} and even \textbf{+10.3 mAP} over \circlenum{1}.
In our experiments, the geometric features performed better than the embedding features (+1.1 mAP).
One possible explanation could be that the geometric features have an explicit meaning and are therefore easier to train than the 5-dimensional embedding space used in experiment \circlenum{3}.
Therefore, for the next experiment in the ablation study and our final model, we make use of the geometric features.
In summary, the quality of the aggregation features has a significant impact.

\parag{Does the graph convolutional network help?} The graph convolutional network (GCN) defined on top of proposals enables higher-order interaction between proposals.
Experiment \circlenum{5} corresponds to the model depicted in \reffig{model} with a 10 layer GCN.
Experiment \circlenum{4} differs from experiment \circlenum{5} in that it does not include the GCN for proposal consolidation.
Adding the GCN results in another improvement of +1.3 mAP.
In total, by incorporating the GCN and replacing NMS with multi-proposal aggregation, we observe an improvement of \textbf{+11.6 mAP} over the same network architecture without those changes.

%% file: results/object_detection_scannet.tex
% !TEX root = ../cvpr20_instance.tex

\begin{table}
\center
    \begin{tabular}{l c c}
        \toprule
        \multicolumn{3}{c}{\textbf{3D Object Detection}}  \\ 
            \midrule
         ScanNetV2 \hspace{2cm}     & mAP@25\% & mAP@50\%  \\
         \midrule
         DSS~\cite{Song16CVPR}      & 15.2  & 6.8  \\
         MRCNN 2D-3D~\cite{He17ICCV}& 17.3  & 10.5 \\
         F-PointNet~\cite{Qi18CVPR} & 19.8  & 10.8 \\
         GSPN~\cite{Li19CVPR}       & 30.6  & 17.7 \\
         3D-SIS~\cite{Hou19CVPR}    & 40.2  & 22.5 \\
         VoteNet~\cite{Qi19ICCV}    & 58.6  & 33.5 \\
         \textbf{\name{} (Ours)}    & \textbf{64.2} & \textbf{49.2}\\
         \bottomrule
    \end{tabular}
    \vspace{-5px}
    \caption{\textbf{3D object detection scores on ScanNetV2}~\cite{Dai17CVPR} validation set.
    We report per-class mean average precision (mAP) with an IoU of 25 \% and 50 \%. The IoU is computed on bounding boxes.
    All other scores are as reported in \cite{Qi19ICCV}.}
    \label{tab:object_detection_scannet}
\end{table}

%% file: results/instance_segmentation_stanford.tex
% !TEX root = ../cvpr20_instance.tex

\begin{table}
	\center
    \vspace{-8px}
    \begin{tabular}{l c c}
        \toprule
        \multicolumn{3}{c}{\textbf{3D Instance Segmentation}}  \\ 
        \midrule
         S3DIS 6-fold CV \hspace{1cm} & mAP@50\% & mAR@50\%  \\
         \midrule
         PartNet~\cite{Mo19CVPR} & 56.4 & 43.4 \\
         ASIS~\cite{Wang19CVPR} & 63.6 & 47.5 \\
         3D-BoNet~\cite{Yang19CVPR} & 65.6 & 47.6 \\
         \textbf{\name~(Ours)} & \textbf{66.7} & \textbf{64.1}\\
         \midrule
         S3DIS Area 5 & mAP@50\% & mAR@50\%  \\
         \midrule
         ASIS~\cite{Wang19CVPR} & 55.3 & 42.4 \\
         3D-BoNet~\cite{Yang19CVPR} & 57.5 & 40.2 \\
         \textbf{\name~(Ours)} & \textbf{63.1} & \textbf{58.0}\\
         \bottomrule
    \end{tabular}
    \vspace{-5px}
    \caption{\textbf{3D instance segmentation scores on S3DIS~\cite{Armeni16CVPR}.} We report scores on Area 5 \emph{(bottom)} and 6-fold cross validation results \emph{(top)}. The metric is mean average precision (mAP) and mean average recall (mAR) at an IoU threshold of 50\%. The IoU is computed on per-point instance masks.}
    \label{tab:inst_stan}
\end{table}

%% file: results/instance_segmentation_scannet.tex
% !TEX root = ../cvpr20_instance.tex

\begin{table}
\setlength{\tabcolsep}{1px}%
\vspace{-8px}
\center
\resizebox{\columnwidth}{!}{

    \begin{tabular}{l c c c c c c c}
        \toprule
        \multicolumn{7}{c}{\textbf{3D Instance Segmentation}}  \\ 
            \midrule
            \multirow{2}{*}{\hspace{3mm}ScanNetV2}  & \multicolumn{3}{c}{Validation Set} & & \multicolumn{3}{c}{Hidden Test Set}\\
           \cmidrule(r){2-4} \cmidrule(r){6-8}
          & mAP & \small{@50\%} & \small{@25\%} & \hspace{3px} & mAP & \small{@50\%} & \small{@25\%}\\
          \midrule
          SGPN~\cite{Wang18CVPR}    		& -     & 11.3  & 22.2 			& & 4.9 & 14.3 & 39.0\\
          3D-BEVIS~\cite{ElichGCPR19}    & -     & -  & - 			& & 11.7 & 24.8 & 40.1\\

          3D-SIS~\cite{Hou19CVPR}   		& -     & 18.7  & 35.7 			& & 16.1 & 38.2 & 55.8\\
          GSPN~\cite{Li19CVPR}      			& 19.3  & 37.8  & 53.4 		& & 15.8 & 30.6 & 54.4\\
          3D-BoNet~\cite{Yang19CVPR} 	& -  		& -  		& - 			& & 25.3 & 48.8 & 68.7\\
          MTML~\cite{Lahoud19ICCV}  		& 20.3  & 40.2  & 55.4 		& & 28.2 & 54.9 & 73.1\\
          
         \textbf{\name{} (Ours)}    & \textbf{35.3}  & \textbf{59.1}  & \textbf{72.4} & & \textbf{35.5} & \textbf{61.1} &  \textbf{73.7} \\
         \bottomrule
    \end{tabular}
}
\vspace{-5px}
 \caption{\textbf{3D instance segmentation scores ScanNetV2}~\cite{Dai17CVPR}. The metric is mean average precision (mAP) at an IoU threshold of 55\%, 50\% and averaged over the range [0.5:0.95:05]. IoU on per-point instance masks.}
\label{tab:inst_scan}
\end{table}

%% file: results/qualitative_results.tex
% !TEX root = ../cvpr20_instance.tex

\definecolor{unlabeled}{rgb}{0., 0., 0.}
\definecolor{wall}{rgb}{0.68235294, 0.78039216, 0.90980392}
\definecolor{floor}{rgb}{0.59607843, 0.8745098 , 0.54117647}
\definecolor{cabinet}{rgb}{0.12156863, 0.46666667, 0.70588235}
\definecolor{bed}{rgb}{1.        , 0.73333333, 0.47058824}
\definecolor{chair}{rgb}{0.7372549 , 0.74117647, 0.13333333}
\definecolor{sofa}{rgb}{0.54901961, 0.3372549 , 0.29411765}
\definecolor{table}{rgb}{1.        , 0.59607843, 0.58823529}
\definecolor{door}{rgb}{0.83921569, 0.15294118, 0.15686275}
\definecolor{window}{rgb}{0.77254902, 0.69019608, 0.83529412}
\definecolor{bookshelf}{rgb}{0.58039216, 0.40392157, 0.74117647}
\definecolor{picture}{rgb}{0.76862745, 0.61176471, 0.58039216}
\definecolor{counter}{rgb}{0.09019608, 0.74509804, 0.81176471}
\definecolor{desk}{rgb}{0.96862745, 0.71372549, 0.82352941}
\definecolor{curtain}{rgb}{0.85882353, 0.85882353, 0.55294118}
\definecolor{refrigerator}{rgb}{1.        , 0.49803922, 0.05490196}
\definecolor{showercurtain}{rgb}{0.61960784, 0.85490196, 0.89803922}
\definecolor{toilet}{rgb}{0.17254902, 0.62745098, 0.17254902}
\definecolor{sink}{rgb}{0.43921569, 0.50196078, 0.56470588}
\definecolor{bathtub}{rgb}{0.89019608, 0.46666667, 0.76078431}
\definecolor{otherfurn}{rgb}{0.32156863, 0.32941176, 0.63921569}

\begin{figure*}
\centering
\begin{tabular}{cccc}
\small\textit{Ground\,Truth\,Instances} &
\small\textit{Predicted\,Instances} &
\small\textit{Predicted\,Object\,Centers} &
\small\textit{\hspace{-10px}Center\,Votes\,\&\,Aggregated\,Proposals} \\
\hspace{0.205\textwidth} &
\hspace{0.208\textwidth} &
\hspace{0.209\textwidth} &
\hspace{0.20\textwidth}
\vspace{-12pt}
\end{tabular}
\includegraphics[width=0.94\linewidth]{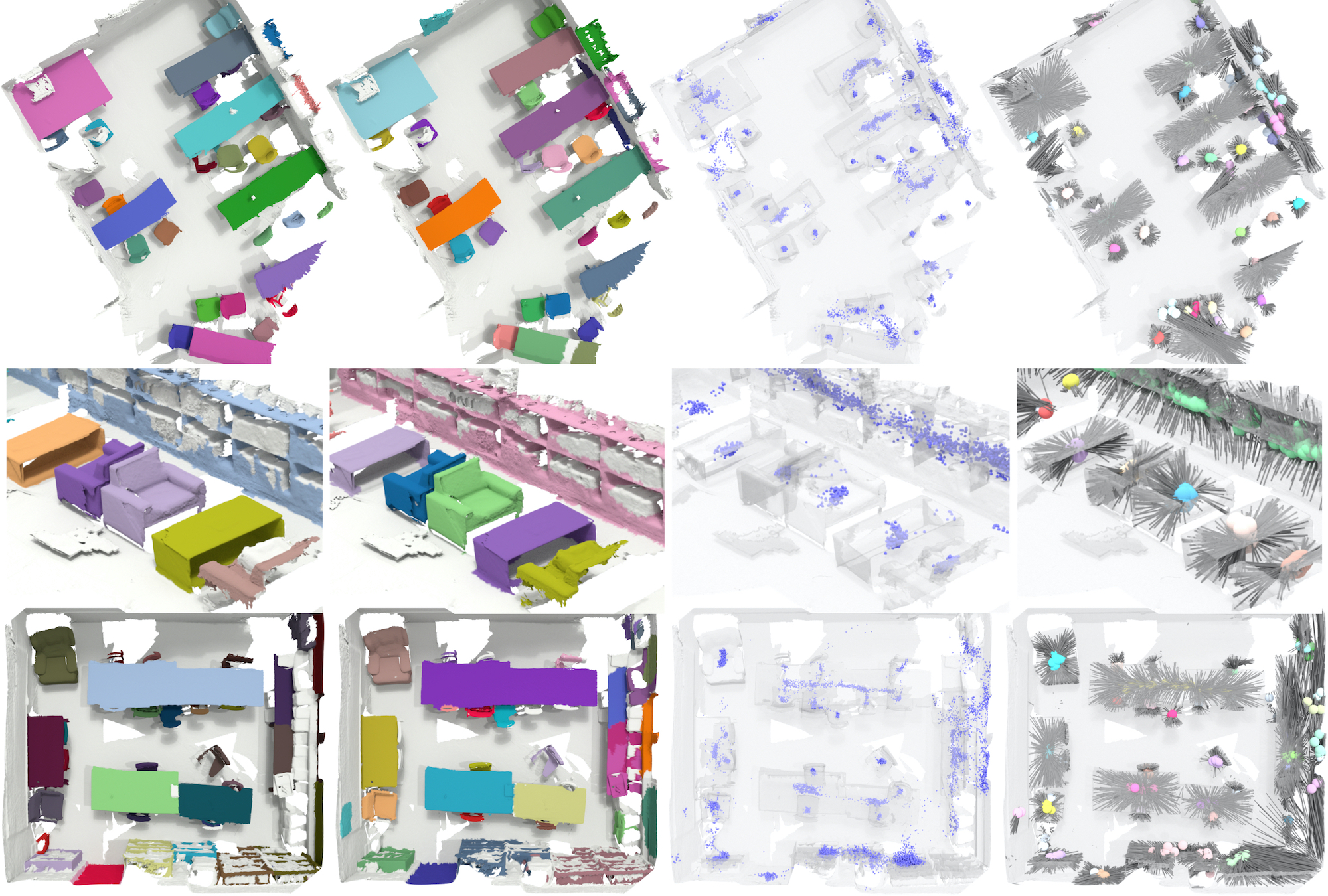}
\vspace{-7pt}
\caption{\textbf{Qualitative results and intermediate steps} on ScanNetV2~\cite{Dai17CVPR}.
First two columns: Our approach properly segments instances of vastly different sizes and makes clear decisions at object boundaries.
Different colors represent separate instances (ground truth and predicted instances are not necessarily the same color).
Third column:
Every point on the surface of an object predicts its object center.
These centers are shown as blue dots.
Fourth column:
Gray segments correspond to votes, they illustrate which point predicted a center. 
Colored spheres represent proposals.
Proposals are obtained by sampling from the predicted object centers. Proposal features are learning from grouped point features that voted for the same object center.
Spheres with the same color show which proposals are grouped together based on these learned proposal features. 
}
\label{fig:qualitative_results}
\vspace{-0.1cm}
\end{figure*}

\begin{figure*}
\centering
\begin{tabular}{cccc}
\small\textit{Ground Truth Instances} &
\small\textit{Predicted Instances} &
\small\textit{Predicted Object Centers} &
\small\textit{Input Point Cloud} \\
\hspace{0.22\textwidth} &
\hspace{0.22\textwidth} &
\hspace{0.22\textwidth} &
\hspace{0.22\textwidth}
\vspace{-14pt}
\end{tabular}
\includegraphics[width=0.94\linewidth]{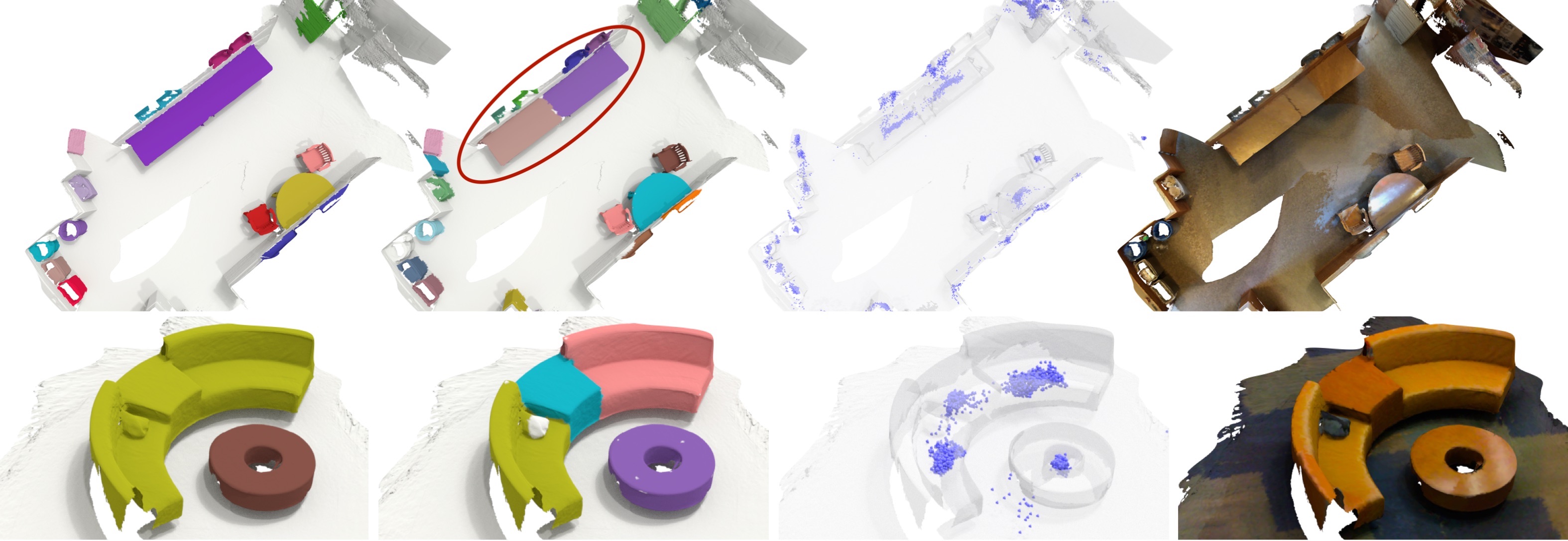}
\vspace{-7pt}
\caption{\textbf{Failure Cases.} We show two failure cases where our method incorrectly separates single instances. However, when comparing them to the input point cloud, they are still plausible predictions.}
\label{fig:failure_cases}
\vspace{0.11cm}
\end{figure*}

%% file: results/instance_segmentation_scannet_class.tex
\begin{table*}
    \resizebox{\textwidth}{!}{
    \begin{tabular}{l|cccccccccccccccccc|c}
    \toprule 
     \textbf{mAP@25} \%& cab & bed & chair & sofa & tabl & door & wind & bkshf & pic & cntr & desk & curt & fridg & showr & toil & sink & bath & ofurn & \textbf{avg}\\ \midrule
     SegCluster~\cite{Hou19CVPR} & 11.8&13.5&18.9&14.6&13.8&11.1&11.5&11.7&0.0&13.7&12.2&12.4&11.2&18.0&19.5&18.9&16.4&12.2 & 13.4\\
     MRCNN~\cite{He17ICCV} & 15.7&15.4&16.4&16.2&14.9&12.5&11.6&11.8&19.5&13.7&14.4&14.7&21.6&18.5&25.0&24.5&24.5&16.9 & 17.1\\
     SGPN~\cite{Wang18CVPR} & 20.7&31.5&31.6&40.6&31.9&16.6&15.3&13.6&0.0&17.4&14.1&22.2&0.0&0.0&72.9&52.4&0.0&18.6 & 22.2 \\ 
    3D-SIS~\cite{Hou19CVPR} & 32.0&66.3&65.3&56.4&29.4&26.7&10.1&16.9&0.0&22.1&35.1&22.6&28.6&37.2&74.9&39.6&57.6&21.1 &35.7 \\
    MTML~\cite{Lahoud19ICCV}& 34.6 & 80.6 & \textbf{87.7} & \textbf{80.3} & 67.4 &45.8 & 47.2 & 45.3 & 19.8 & 9.7 & 49.9 & 54.2 & 44.1 & 74.9 & 98.0 & 44.5 & 79.4 & 33.5 & 55.4\\
\midrule
\name{} (Ours) & \textbf{69.9} & \textbf{83.4} & 87.6&76.1&\textbf{74.8}& \textbf{56.6} & \textbf{62.2} & \textbf{78.3} & \textbf{48.0} & \textbf{62.5} & \textbf{69.2} & \textbf{66.0} & \textbf{61.4} & \textbf{93.1} & \textbf{99.2} & \textbf{75.2} &\textbf{90.3} & \textbf{48.6} & \textbf{72.4} \\ 
\bottomrule
    \end{tabular}
    }
    \caption{\textbf{Per class 3D instance segmentation} on ScanNetV2~\cite{Dai17CVPR} validation set with mAP@25\% on 18 classes. Our method outperforms all other methods on all classes except for \emph{chair} and \emph{sofa}.}
    \label{tab:instance_segmentation_scannet_class_map25}
\end{table*}

\begin{table*}
    \resizebox{\textwidth}{!}{
    \begin{tabular}{l|cccccccccccccccccc|c}
    \toprule
    \textbf{mAP@50\%} & cab & bed & chair & sofa & tabl & door & wind & bkshf & pic & cntr & desk & curt & fridg & showr & toil & sink & bath & ofurn & \textbf{avg}\\
    \midrule
    SegCluster~\cite{Hou19CVPR}&10.4&11.9&15.5&12.8&12.4&10.1&10.1&10.3&0.0&11.7&10.4&11.4&0.0&13.9&17.2&11.5&14.2&10.5&10.8\\
    MRCNN~\cite{He17ICCV}&11.2&10.6&10.6&11.4&10.8&10.3&0.0&0.0&11.1&10.1&0.0&10.0&12.8&0.0&18.9&13.1&11.8&11.6&9.1\\
    SGPN~\cite{Wang18CVPR}&10.1&16.4&20.2&20.7&14.7&11.1&11.1&0.0&0.0&10.0&10.3&12.8&0.0&0.0&48.7&16.5&0.0&0.0&11.3\\
   3D-SIS~\cite{Hou19CVPR}&19.7&37.7&40.5&31.9&15.9&18.1&0.0&11.0&0.0&0.0&10.5&11.1&18.5&24.0&45.8&15.8&23.5&12.9&18.7\\
   MTML~\cite{Lahoud19ICCV}&14.5&54.0&79.2&48.8&42.7&32.4&32.7&21.9&10.9&0.8&14.2&39.9&42.1&64.3&96.5&36.4&70.8&21.5&40.2\\
   \midrule
   \name{} (Ours) &\textbf{51.9}&\textbf{72.2}&\textbf{83.8}&\textbf{66.8}&\textbf{63.0}&\textbf{43.0}&\textbf{44.5}&\textbf{58.4}&\textbf{38.8}&\textbf{31.1}&\textbf{43.2}&
   \textbf{47.7}&\textbf{61.4}&\textbf{80.6}&\textbf{99.2}&\textbf{50.6}&\textbf{87.1}&\textbf{40.3}&\textbf{59.1}\\
   \bottomrule
   \end{tabular}
    }
    \caption{\textbf{Per class 3D instance segmentation} on ScanNetV2~\cite{Dai17CVPR} validation set with mAP@50\% on 18 classes. Our method outperforms all other methods on all classes.}
    \label{tab:instance_segmentation_scannet_class_map50}
\end{table*}

%% file: results/ablation_study.tex
% !TEX root = ../cvpr20_instance.tex

\begin{table}[b]
\setlength{\tabcolsep}{4px}%
\resizebox{\columnwidth}{!}{
    \begin{tabular}{l l}
        \toprule
        \multicolumn{2}{c}{\textbf{Ablation Study}}  \\ 
        \midrule
		3D Instance Segmentation (ScanNetV2 val.) & mAP\small{@50\%}\\
         \midrule
         \circlenum{1} Proposals + NMS & 47.5 \\
         \circlenum{2} Agg. Props. (proposal positions) & 52.4 \textcolor{darkgreen}{\small{(+\,4.9)}}\\
         \circlenum{3} Agg. Props. (embedding features) & 56.7 \textcolor{darkgreen}{\small{(+\,9.2)}}\\
         \circlenum{4} Agg. Props. (geometric features) & 57.8 \textcolor{darkgreen}{\small{(+\,10.3)}}\\
         \circlenum{5} Agg. Props. (geometric features + GCN) & \textbf{59.1} \textcolor{darkgreen}{\small{(+\,11.6)}}\\
         \bottomrule
    \end{tabular}
    }
     \caption{\textbf{Ablation study.}
     In \refsec{ablation} we discuss the results in detail.
     Scores are instance segmentation results on the ScanNetV2~\cite{Dai17CVPR} validation set
     and absolute improvements in mAP (in green) relative to the baseline \circlenum{1}.}
    \label{tab:ablation_scannet}
\end{table}

%% file: results/instance_segmentation_stanford_class.tex
\begin{table*}[t]
\begin{center}
\resizebox{\textwidth}{!}{
\begin{tabular}{l|l|ccccccccccccc|c}
\toprule 
&S3DIS 6-fold CV & ceil. & floor & walls & beam & colm, & wind. & door & table & chair & sofa & bookc. & board & clut. & \textbf{mean}\\
\midrule
 \multirow{2}{*}{mAP@0.5} 
 & 3D-BoNet~\cite{Yang19CVPR} & 88.5 & 89.9 & \textbf{64.9} & 42.3 & 48.0 & \textbf{93.0} & 66.8 & \textbf{55.4} & 72.0 & 49.7 & \textbf{58.3} & \textbf{80.7} & \textbf{47.6} & 65.6\\
& \name~(Ours) & \textbf{95.5} & \textbf{99.5} & 59.0 & \textbf{44.6} & \textbf{57.7} & 89.0 & \textbf{78.7} & 34.5 & \textbf{83.6} & \textbf{55.9} & 51.6 & 71.0 & 46.3 & \textbf{66.7} \\
\midrule
 \multirow{2}{*}{mAR@0.5} 
 & 3D-BoNet~\cite{Yang19CVPR} & 61.8 & 74.6 & 50.0 & 42.2 & 27.2 & 62.4 & 58.5 & \textbf{48.6} & 64.9 & 28.8 & 28.4 & 46.5 & 28.6 & 46.7\\
 & \name~(Ours)                 & \textbf{68.4} & \textbf{96.2} & \textbf{51.9} & \textbf{58.8} & \textbf{77.6} & \textbf{79.8} & \textbf{69.5} & 32.8 & \textbf{75.2} & \textbf{71.1} & \textbf{46.2} & \textbf{68.2} & \textbf{38.2} & \textbf{64.1}\\
\bottomrule
\end{tabular}
}
\caption{\textbf{Per class 3D instance segmentation scores on S3DIS}~\cite{Armeni16CVPR}.
We report per-class mean average precision (mAP) and recall (mAR) with an IoU of 50\,\%.
3D-BoNet are up-to-date numbers provided by the original authors. Our method detects significantly more objects (+\,17.4\,mAR) and it is even able to do so with a higher precision (+\,1.1\,mAP).}
\label{tab:instance_segmentation_stanford_class}
\end{center}
\end{table*}

%% file: sections/5_conclusion.tex
% !TEX root = ../cvpr20_instance.tex

\section{Conclusion}
In this work, we introduced \name{}, a new method for 3D semantic instance segmentation.
Our core idea is to combine the benefits of both top-down and bottom-up object detection strategies.
That is, we first produce a number of proposals using an object-centric voting scheme based on a sparse volumetric backbone.
Each object may receive multiple proposals, which makes our method robust to potential outliers in the object proposal stage.
However, at the same time we obtain only a handful of proposals such that clustering them is computationally inexpensive.
To address this, we first allow higher-order feature interactions between proposals via a graph convolutional network. 
We then aggregate proposals based on graph relationship results and proposal feature similarities. 
We show that graph convolutions help to achieve high evaluation scores,
although, the largest improvement originates from our multi proposal aggregation strategy.
Our combined approach achieves state-of-the-art instance segmentation and object detection results on the popular ScanNetV2 and S3DIS datasets, thus validating our algorithm design.

Overall, we believe that multi proposal aggregation is a promising direction for object detection, in particular in the 3D domain. 
However, there still remain many interesting future avenues, for instance, how to combine detection with tracking in semi-dynamic sequences. 
We see a variety of interesting ideas, where proposals could be distributed in 4D space and accumulated along the time-space axis.

\parag{Acknowledgements.} 
We would like to thank Theodora Kontogianni, Jonas Schult, Jonathon Luiten, Mats Steinweg, Ali Athar, Dan Jia and Sabarinath Mahadevan for helpful feedback
as well as Angela Dai for help with the video.
This work was funded by the ERC Consolidator Grant DeeViSe (ERC-2017-COG-773161) and the ERC Starting Grant Scan2CAD (804724).